\documentclass[final]{cvpr}
\usepackage{times}
\usepackage{epsfig}
\usepackage{graphicx}
\usepackage{amsmath}
\usepackage{amssymb}
\usepackage{color, colortbl}
\usepackage{multirow}
\usepackage{enumitem}
\usepackage{amsmath}
\usepackage{makecell}
\usepackage{hhline}
\usepackage{cellspace}
\usepackage{bbding}
\usepackage{bm}
\usepackage{nicefrac}       % compact symbols for 1/2, etc.
\usepackage{microtype}      % microtypography
\usepackage{verbatim}
\usepackage{color}
\usepackage{float}
\usepackage{enumitem}
\usepackage{booktabs}
\usepackage{tabulary,overpic,xcolor}
\definecolor{mygray}{gray}{.92}
\definecolor{LightCyan}{rgb}{0.88,1,1}
\usepackage[caption=false]{subfig}
\usepackage[pagebackref=true,breaklinks=true,letterpaper=true,colorlinks,bookmarks=false]{hyperref}
\def\cvprPaperID{511} % *** Enter the CVPR Paper ID here
\def\confYear{CVPR 2021}
\newcolumntype{x}[1]{>{\centering\arraybackslash}p{#1pt}}
\newlength\savewidth\newcommand\shline{\noalign{\global\savewidth\arrayrulewidth
		\global\arrayrulewidth 1pt}\hline\noalign{\global\arrayrulewidth\savewidth}}
\newcommand{\tablestyle}[2]{\setlength{\tabcolsep}{#1}\renewcommand{\arraystretch}{#2}\centering\footnotesize}
\newcommand{\myparagraph}[1]{{\vspace{0.5em} \noindent \bf #1}}

\usepackage[utf8]{inputenc}
\usepackage{cleveref}
\crefname{section}{§}{§§}
\Crefname{section}{§}{§§}

\begin{document}

%%%%%%%%% TITLE
\title{Track to Detect and Segment: An Online Multi-Object Tracker}
\author{Jialian Wu$^{1}$, Jiale Cao$^{2}$, Liangchen Song$^{1}$, Yu Wang$^{3}$, Ming Yang$^{3}$, Junsong Yuan$^{1}$\\\\
	$^1$SUNY Buffalo~~~~~~~~~$^2$TJU~~~~~~~~~$^3$Horizon Robotics
}

\maketitle
%%%%%%%%% ABSTRACT
\begin{abstract}
Most online multi-object trackers perform object detection stand-alone in a neural net without any input from tracking. In this paper, we present a new online joint detection and tracking model, TraDeS (TRAck to DEtect and Segment), exploiting tracking clues to assist detection end-to-end. TraDeS infers object tracking offset by a cost volume, which is used to propagate previous object features for improving current object detection and segmentation. Effectiveness and superiority of TraDeS are shown on 4 datasets, including MOT (2D tracking), nuScenes (3D tracking), MOTS and Youtube-VIS (instance segmentation tracking). Project page: \url{https://jialianwu.com/projects/TraDeS.html}.
\end{abstract}

%%%%%%%%% BODY TEXT\left( 
\section{Introduction}
\begin{figure}
	\centering
	\includegraphics[width=1\linewidth]{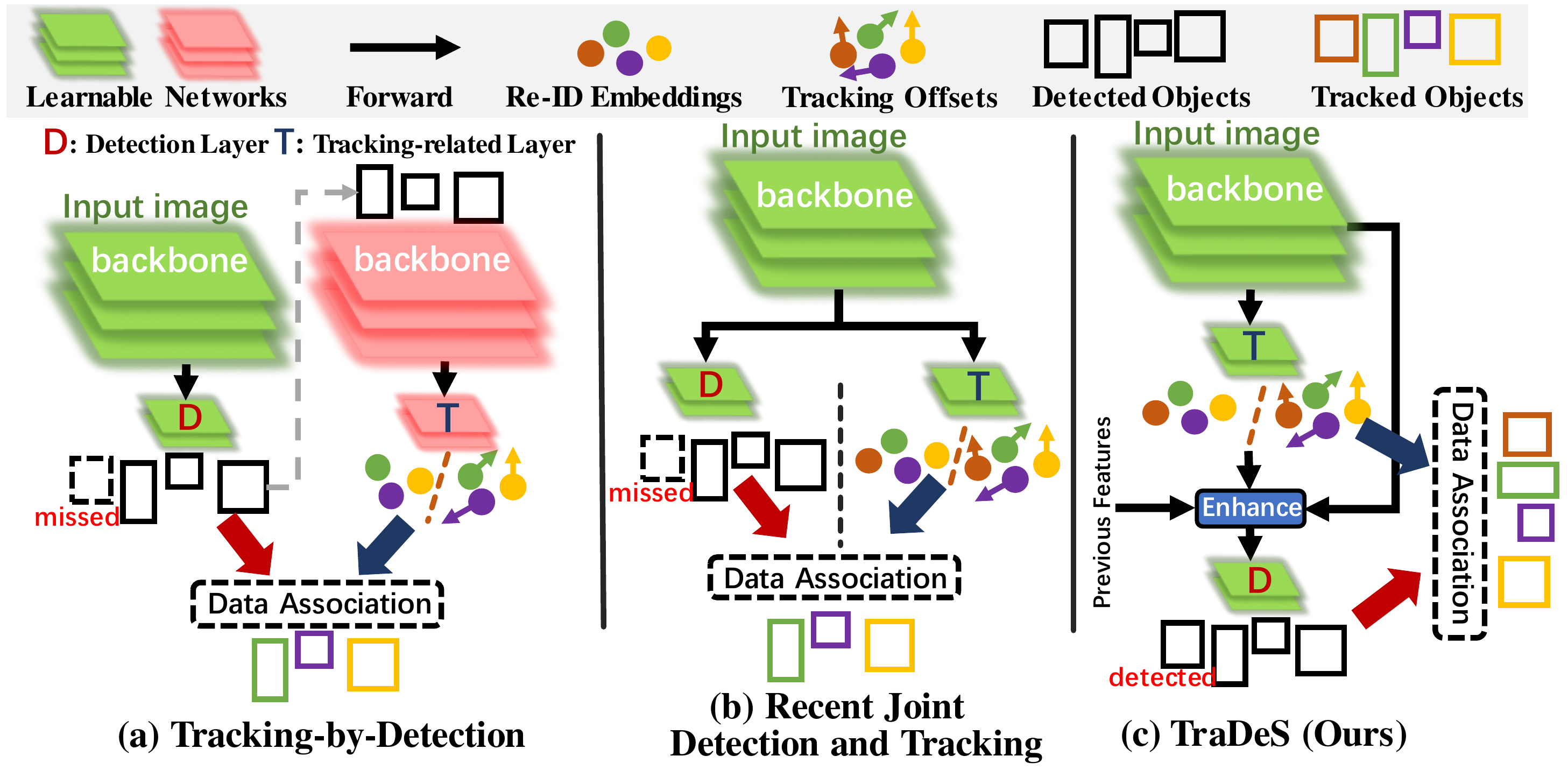}
	\caption{\textbf{Comparison of different online MOT pipelines.} Our method follows the joint detection and tracking (\emph{JDT}) paradigm. Different from most \emph{JDT} methods, the proposed TraDeS tracker deeply couples tracking and detection within an end-to-end and unified framework, where the motion clue from tracking is exploited to enhance detection or segmentation (omitted in the figure).}
	\label{fig:fig1}
	\vspace{-2mm}
\end{figure}

Advanced online multi-object tracking methods follow two major paradigms: tracking-by-detection~\cite{bewley2016simple,tang2017multiple,milan1603mot16,xu2019spatial,Porzi_2020_CVPR,wojke2017simple} and joint detection and tracking~\cite{meinhardt2021trackformer,CenterTrack,tracktor,CTacker,wang2019towards,lu2020retinatrack,wang2020combining,wang2020joint}. The tracking-by-detection (\emph{TBD}) paradigm treats detection and tracking as two independent tasks (Fig.~\ref{fig:fig1} (a)). It usually applies an off-the-shelf object detector to produce detections and employs another separate network for data association. The \emph{TBD} system is inefficient and not optimized end-to-end due to the two-stage processing. To address this problem, recent solutions favor a joint detection and tracking (\emph{JDT}) paradigm that simultaneously performs detection and tracking in a single forward-pass (Fig.~\ref{fig:fig1} (b)).

The \emph{JDT} methods, however, are confronted with two issues: \textbf{(i)} Although in most \emph{JDT} works~\cite{CTacker,wang2019towards,lu2020retinatrack,wu2020temporal} the backbone network is shared, detection is usually performed standalone without exploring tracking cues. We argue that detection is the cornerstone for a stable and consistent tracklet, and in turn tracking cues shall help detection, especially in tough scenarios like partial occlusion and motion blur.  \textbf{(ii)} As studied by~\cite{chen2018person} and our experiment (Tab.~\ref{tab:ablation2}), common re-ID tracking loss~\cite{wang2019towards,lu2020retinatrack,schroff2015facenet,xiao2017joint} is not that compatible with detection loss in jointly training a single backbone network, which could even hurt detection performance to some extent. The reason is that re-ID focuses on intra-class variance, but detection aims to enlarge inter-class difference and minimize intra-class variance. 

In this paper, we propose a new online joint detection and tracking model, coined as TraDeS (TRAck to DEtect and Segment). In TraDeS, each point on the feature map represents either an object center or a background region, similar as in CenterNet~\cite{zhou2019objects}. TraDeS addresses the above two issues by tightly incorporating tracking into detection as well as a dedicatedly designed re-ID learning scheme. Specifically, we propose a cost volume based association (CVA) module and a motion-guided feature warper (MFW) module, respectively. The CVA extracts point-wise re-ID embedding features by the backbone to construct a cost volume that stores matching similarities between the embedding pairs in two frames. Then, we infer the tracking offsets from the cost volume, which are the spatio-temporal displacements of all the points, \emph{i.e.,} potential object centers, in two frames. The tracking offsets together with the embeddings are utilized to conduct a simple two-round long-term data association. Afterwards, the MFW takes the tracking offsets as motion cues to propagate object features from the previous frames to the current one. Finally, the propagated feature and the current feature are aggregated to derive detection and segmentation.

In the CVA module, the cost volume is employed to supervise the re-ID embedding, where different object classes and background regions are implicitly taken into account. This is being said, our re-ID objective involves the inter-class variance. This way not only learns an effective embedding as common re-ID loss~\cite{wang2019towards, lu2020retinatrack, schroff2015facenet, xiao2017joint}, but also is well compatible with the detection loss and does not hurt detection performance as shown in Tab.~\ref{tab:ablation2}. Moreover, because the tracking offset is predicted based on appearance embedding similarities, it can match an object with very large motion or in low frame rate as shown in Fig.~\ref{fig:ablation1}, or even accurately track objects in different datasets with unseen large motion as shown in Fig.~\ref{fig:ablation4}. Thus, the predicted tracking offset of an object can serve as a robust motion clue to guide our feature propagation in the MFW module. The occluded and blurred objects in the current frame may be legible in early frames, so the propagated features from previous frames may support the current feature to recover potentially missed objects by our MFW module.

In summary, we propose a novel online multi-object tracker, TraDeS, that deeply integrates tracking cues to assist detection in an end-to-end framework and in return benefits tracking as shown in Fig.~\ref{fig:fig1} (c). TraDeS is a general tracker, which is readily extended to instance segmentation tracking by adding a simple instance segmentation branch. Extensive experiments are conducted on 4 datasets, \emph{i.e.}, MOT, nuScenes, MOTS, and Youtube-VIS datasets, across 3 tasks including 2D object tracking, 3D object tracking, and instance segmentation tracking. TraDeS achieves state-of-the-art performance with an efficient inference time as shown in~\cref{subsec:sota}. Additionally, thorough ablation studies are performed to demonstrate the effectiveness of our approach as shown in~\cref{subsec:ablation_studies}.

\section{Related Work}
\myparagraph{Tracking-by-Detection.} MOT was dominated by the tracking-by-detection (\emph{TBD}) paradigm over the past years~\cite{yin2020unified,braso2020learning,zhu2018online,xu2019spatial,schulter2017deep,bewley2016simple,tang2017multiple,Weng_2020_CVPR,xu2020segment}. Within this framework, an off-the-shelf object detector~\cite{ren2015faster,felzenszwalb2009object} is first applied to generate detection boxes for each frame. Then, a separate re-ID model~\cite{tracktor,wojke2017simple} is used to extract appearance features for those detected boxes. To build tracklets, one simple solution is to directly compute appearance and motion affinities with a motion model, \emph{e.g.,} Kalman filter, and then solve data association by a matching algorithm. Some other efforts~\cite{braso2020learning,wen2014multiple,kim2015multiple} formulate data association as a graph optimization problem by treating each detection as a graph node. However, \emph{TBD} methods conduct detection and tracking separately, hence are usually computationally expensive. Instead, our approach integrates tracking cues into detection and efficiently performs detection and tracking in an end-to-end fashion.

\myparagraph{Joint Detection and Tracking.} Recently joint detection and tracking (\emph{JDT}) paradigm has raised increasing attention due to its efficient and unified framework. One common way~\cite{CenterTrack,wang2019towards,lu2020retinatrack,tracktor,zhang2018integrated,zhang2020fairmot} is to build a tracking-related branch upon an object detector to predict either object tracking offsets or re-ID embeddings for data association. Alternatively, transformer is exploited to match tracklets~\cite{sun2020transtrack,meinhardt2021trackformer}. CTracker~\cite{CTacker} constructs tracklets by chaining paired boxes in every two frames. TubeTK~\cite{tubetk} directly predicts a box tube as a tracklet in an offline manner. Most \emph{JDT} methods, however, are confronted with two issues: First, detection is still separately predicted without the help from tracking. Second, the re-ID loss has a different objective from that of detection loss in joint training. In contrast, our TraDeS tracker addresses these two problems by tightly incorporating tracking cues into detection and designing a novel re-ID embedding learning scheme.

\myparagraph{Tracking-guided Video Object Detection.} In video object detection, a few attempts~\cite{feichtenhofer2017detect,zhang2018integrated} exploit tracking results to reweight the detection scores generated by an initial detector. Although these works strive to help detection by tracking, they have two drawbacks: First, tracking is leveraged to help detection only at the post-processing stage. Detections are still predicted by a standalone object detector, so detection and tracking are separately optimized. Thus, the final detection scores may heavily rely on the tracking quality. Second, a hand-crafted reweighting scheme requires manual tune-up for a specific detector and tracker. Our approach differs from these post-processing methods because our detection is learned conditioned on tracking results, without a complex reweighting scheme. Therefore, detection tends to be robust \emph{w.r.t.} tracking quality. 

\myparagraph{Cost Volume.} The cost volume technique has been successfully applied in depth estimation~\cite{collins1996space,yang2020cost,im2019dpsnet} and optical flow estimation~\cite{sun2018pwc,chen2016full,xu2017accurate} for associating pixels between two frames. This motivates us to extend cost volume to a multi-object tracker, which will be demonstrated to be effective in learning re-ID embeddings and inferring tracking offsets in this paper. Our approach may inspire future works using cost volume in tracking or re-identification.

\begin{figure*}
	\centering
	\includegraphics[width=1\linewidth]{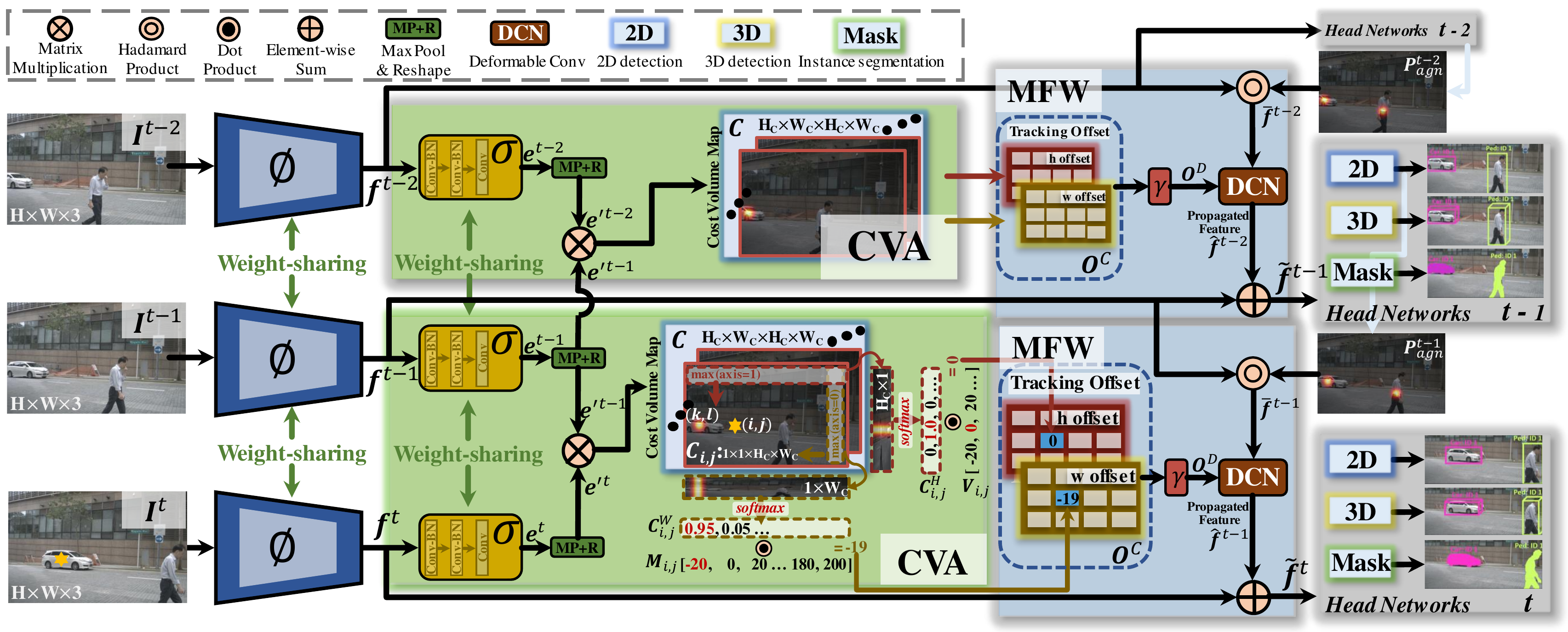}
	\caption{\textbf{Overview of TraDeS.} TraDeS may propagate features from multiple previous frames for object feature enhancement (\emph{i.e.,} $T>1$), which is not shown in the above figure for simplicity.}
	\label{fig:overview}
	\vspace{-2mm}
\end{figure*}

\section{Preliminaries}
\label{sec:preliminaries}
The proposed TraDeS is built upon the point-based object detector CenterNet~\cite{zhou2019objects}. CenterNet takes an image $\bm{I} \in\mathbb{R}^{H\times W\times 3}$ as input and produces the base feature $\bm{f}=\phi(\bm{I})$ via the backbone network $\phi(\cdot)$, where $\bm{f} \in\mathbb{R}^{H_{F}\times W_{F}\times 64}$, $H_{F} = \frac{H}{4}$, and $W_{F} = \frac{W}{4}$. A set of head convolutional branches are then constructed on $\bm{f}$ to yield a class-wise center heatmap $\bm{P}\in\mathbb{R}^{H_{F}\times W_{F}\times N_{cls}}$ and task-specific prediction maps, such as 2D object size map and 3D object size map, etc. $N_{cls}$ is the number of classes. CenterNet detects objects by their center points (peaks on $\bm{P}$) and the corresponding task-specific predictions from the peak positions. 

%\myparagraph{Baseline Tracker:}
Similar to \cite{CenterTrack}, we build a baseline tracker by adding an extra head branch on CenterNet that predicts a tracking offset map $\bm{O}^{B}\in\mathbb{R}^{H_{F}\times W_{F}\times 2}$ for data association. $\bm{O}^{B}$ computes spatio-temporal displacements from all points at time $t$ to the corresponding points at a previous time $t-\tau$.

\section{TraDeS Tracker}
\vspace{-1mm}
\label{sec:trades}
\myparagraph{Our Idea:} Most previous joint detection and tracking methods perform a standalone detection without explicit input from tracking. In contrast, our aim is to integrate tracking cues into detection end-to-end, so as to improve detection for tough scenarios, which in return benefit tracking. To this end, we propose a Cost Volume based Association (CVA:~\cref{subsec:CVA}) module for learning re-ID embeddings and deriving object motions,  and a Motion-guided Feature Warper (MFW:~\cref{subsec:MFW}) module for leveraging tracking cues from the CVA to propagate and enhance object features.

\vspace{-1mm}
\subsection{Cost Volume based Association}
\vspace{-2mm}
\label{subsec:CVA}
\myparagraph{Cost Volume:} Given two base features $\bm{f}^{t}$ and $\bm{f}^{t-\tau}$ from $\bm{I}^{t}$ and $\bm{I}^{t-\tau}$, we extract their re-ID embedding features by the embedding network $\sigma(\cdot)$, \emph{i.e.,} $\bm{e}^{t} = \sigma(\bm{f}^{t})\in \mathbb{R}^{H_{F}\times W_{F}\times 128}$, where  $\sigma(\cdot)$ consists of three convolution layers. We utilize the extracted embeddings to construct a cost volume which stores dense matching similarities between one point and its corresponding point in two frames. To efficiently compute the cost volume, we first downsample the embeddings by a factor of $2$ and obtain $\bm{e}^{\prime} \in \mathbb{R}^{H_{C}\times W_{C}\times 128}$, where $H_{C} = \frac{H_{F}}{2}$ and $W_{C} = \frac{W_{F}}{2}$. Let us denote by $\bm{C} \in \mathbb{R}^{H_{C}\times W_{C}\times H_{C}\times W_{C}}$ the 4-dimensional cost volume for $\bm{I}^{t}$ and $\bm{I}^{t-\tau}$, which is computed by a single matrix multiplication of $\bm{e}^{\prime t}$ and $\bm{e}^{\prime t-\tau}$. Specifically, each element of $\bm{C}$ is calculated as:
\vspace{-2mm}
\begin{equation}
\label{eqn:costvolume_map}
C_{i,j,k,l} = \bm{e}^{\prime t }_{i,j} \bm{e}^{\prime t-\tau \top}_{k,l},
\vspace{-1mm}
\end{equation}
where $C_{i,j,k,l}$ represents the embedding similarity between point $(i,j)$ at time $t$ and point $(k,l)$ at time $t-\tau$. Here, a point refers to an entry on the feature map $\bm{f}$ or $\bm{e}^{\prime}$. 

\myparagraph{Tracking Offset:} Based on the cost volume $\bm{C}$, we calculate
a tracking offset matrix $\bm{O} \in \mathbb{R}^{H_{C}\times W_{C}\times 2}$, which stores the spatio-temporal displacements for all points at time $t$ to their corresponding points at time $t-\tau$. For illustration, we show the estimation procedure for $\bm{O}_{i,j} \in \mathbb{R}^{2}$ below.

As shown in Fig.~\ref{fig:overview}, for an object $x$ centered at point $(i,j)$ at time $t$, we can fetch from $\bm{C}$ its corresponding two-dimensional cost volume map $\bm{C}_{i,j} \in \mathbb{R}^{H_{C}\times W_{C}}$. $\bm{C}_{i,j}$ stores the matching similarities among object $x$ and all points at time $t-\tau$. Using $\bm{C}_{i,j}$, $\bm{O}_{i,j} \in \mathbb{R}^{2}$ is estimated by two steps: 
\textbf{\emph{Step (i)}} $\bm{C}_{i,j}$ is first max pooled by $H_{C}\times1$ and $1\times W_{C}$ kernels, respectively, and then normalized by a softmax function\footnote{We add a temperature of $5$ into the softmax, such that the softmax output values are more discriminative.}, which results in $\bm{C}^{W}_{i,j} \in [0,1]^{1\times W_{C}}$ and $\bm{C}^{H}_{i,j} \in [0,1]^{H_{C}\times 1}$. $\bm{C}^{W}_{i,j}$ and $\bm{C}^{H}_{i,j}$ consists of the likelihoods that object $x$ appears on specified horizontal and vertical positions at time $t-\tau$, respectively. For example, $C^{W}_{i,j,l}$ is the likelihood that object $x$ appears at the position $(\ast,l)$ at time $t-\tau$. 
\textbf{\emph{Step (ii)}} Since $\bm{C}^{W}_{i,j}$ and $\bm{C}^{H}_{i,j}$ have provided the likelihoods that object $x$ appears on specified positions at $t-\tau$. To obtain the final offsets, we predefine two offset templates for horizontal and vertical directions, respectively, indicating the actual offset values when $x$ appears on those positions. Let $\bm{M}_{i,j}\in \mathbb{R}^{1\times W_{C}}$ and $\bm{V}_{i,j} \in \mathbb{R}^{H_{C} \times 1}$ denote the horizontal and vertical offset templates for object $x$, respectively, which are computed by:
\vspace{-2mm}
\begin{equation}
\label{eqn:offs_template}
\begin{cases}
M_{i,j,l} = (l - j) \times s
&1\leq l \leq W_{C}\\
V_{i,j,k} = (k - i) \times s
&1\leq k \leq H_{C}
\end{cases}
,
\end{equation}
where $s$ is the feature stride of $\bm{e}^{\prime}$ \emph{w.r.t.} the input image, which is $8$ in our case. $M_{i,j,l}$ refers to the horizontal offset when object $x$ appears at the position $(\ast,l)$ at time $t-\tau$. 
The final tracking offset can be inferred by the dot product between the likelihoods and actual offset values as:
\vspace{-1mm}
\begin{equation}
\label{eqn:tracking_offset}
\bm{O}_{i,j} = [\bm{C}^{H\top}_{i,j}\bm{V}_{i,j},\bm{C}^{W}_{i,j}\bm{M}_{i,j}^{\top}]^{\top}.
\end{equation}
Because $\bm{O}$ is of $H_{C} \times W_{C}$, we upsample it with a factor of $2$ and obtain $\bm{O}^{C}  \in \mathbb{R}^{H_{F} \times  W_{F} \times 2} $ that serves as motion cues for the MFW and is used for our data association.

\myparagraph{Training:} Since $\sigma(\cdot)$ is the only learnable part in the CVA module, the training objective of CVA is to learn an effective re-ID embedding $\bm{e}$. To supervise $\bm{e}$, we enforce the supervision on the cost volume rather than directly on $\bm{e}$ like other common re-ID losses. Let us first denote $Y_{ijkl}=1$ when an object at location $(i,j)$ at current time $t$ appears at location $(k,l)$ at previous time $t-\tau$; otherwise $Y_{ijkl}=0$. Then, the training loss for CVA is calculated by the logistic regression in the form of the focal loss~\cite{lin2017focal} as:
\vspace{-2mm}
\begin{equation}
\label{eqn:cva_loss}
L_{CVA}\! = \! \frac{-1}{\sum_{ijkl}Y_{ijkl}} \sum_{ijkl}
\begin{cases}
\begin{array}{c}
\!\!\!\! \alpha_{1}\log(C^{W}_{i,j,l})\\ \!\!\!\! +\alpha_{2}\log(C^{H}_{i,j,k})
\end{array} & \!\!\!\!\! \text{if}\ Y_{ijkl}=1\\
\quad \quad\quad 0
& \!\!\!\!\!\text{otherwise}
\end{cases}
,
\end{equation}
where $\alpha_{1}=(1 - C^{W}_{i,j,l})^{\beta}$ and $\alpha_{2}=(1 - C^{H}_{i,j,k})^{\beta}$. $\beta$ is the focal loss hyper-parameter. 
Since $C^{W}_{i,j,l}$ and $C^{H}_{i,j,k}$ are computed by softmax, they involve the embedding similarities not only between points $(i,j)$ and $(k,l)$ but also among point $(i,j)$ and all other points in the previous frame. This is being said, while $C^{W}_{i,j,l}$ and $C^{H}_{i,j,k}$ being optimized to approach $1$, it enforces an object to not only approach itself in the previous frame, but also \emph{repel} other objects and background regions.

\myparagraph{The CVA Characteristics:} \textbf{\emph{(i)}} Common re-ID loss only emphasizes intra-class variance, which may degrade detection performance. In contrast, our $L_{CVA}$ in Eq.~\ref{eqn:cva_loss} not only emphasizes intra-class variance but also forces inter-class difference when learning embedding. We find such a treatment is more compatible with detection loss and learns effective embedding without hurting detection as evidenced in Tab.~\ref{tab:ablation2}. \textbf{\emph{(ii)}} Because the tracking offset is predicted based on appearance embedding similarities, it can track objects under a wide range of motion and low frame rate as shown in Fig.~\ref{fig:ablation1} and Fig.~\ref{fig:final_results}, or even accurately track objects in different datasets with unseen large motion in training set as shown in Fig.~\ref{fig:ablation4}. The predicted tracking offset can therefore serve as a robust motion cue to guide our feature propagation as in Tab.~\ref{tab:ablation3}. \textbf{\emph{(iii)}} Compared to~\cite{wang2019towards,lu2020retinatrack} and CenterTrack~\cite{CenterTrack} that only predict either embedding or tracking offset for data association, the CVA produces both embedding and tracking offset that are used for long-term data association (\cref{subsec:tracklet_generation}) and serve as motion cues for the MFW (\cref{subsec:MFW}).

\subsection{Motion-guided Feature Warper}
\label{subsec:MFW}
The MFW aims to take the predicted tracking offset $\bm{O}^{C}$ as motion clues to warp and propagate $\bm{f}^{t-\tau}$ to the current time so as to compensate and enhance $\bm{f}^{t}$. To achieve this goal, we perform an efficient temporal propagation via a single deformable convolution~\cite{dai2017deformable}, which has been used for temporally aligning features in previous works~\cite{bertasius2018object,bertasius2019learning,deng2020single}. Then, we enhance $\bm{f}^{t}$ by aggregating the propagated feature.

\myparagraph{Temporal Propagation:} 
To propagate feature maps, the deformable convolution (DCN) takes a spatio-temporal offset map and a previous feature as input and outputs a propagated feature, in which we estimate the input offset based on the $\bm{O}^{C}$ from the CVA module. Let us denote $\bm{O}^{D} \in \mathbb{R}^{H_{F} \times W_{F} \times2K^{2}}$ as the input two-directional offset for DCN, where $K=3$ is the kernel width or height of DCN. To generate $\bm{O}^{D}$, we pass $\bm{O}^{C}$ through a $3\times3$ convolution $\gamma(\cdot)$. We optionally incorporate the residual feature of $\bm{f}^{t}-\bm{f}^{t-\tau}$ as the input of $\gamma(\cdot)$ to provide more motion clues. Since our detection and segmentation are mainly based on object center features, instead of directly warping $\bm{f}^{t-\tau}$, we propagate a center attentive feature $\bm{\bar{f}}^{t-\tau}\in \mathbb{R}^{H_{F} \times W_{F} \times 64}$ from previous time. $\bm{\bar{f}}^{t-\tau}$ is computed as:
\vspace{-1mm}
\begin{equation}
\label{eqn:feat_mask}
\bm{\bar{f}}^{t-\tau}_{q}=\bm{f}^{t-\tau}_{q}\circ\bm{P}_{agn}^{t-\tau}, \quad q=1,2,...,64,
\end{equation}
where $q$ is the channel index,  $\circ$ is the Hadamard product, and $\bm{P}_{agn}^{t-\tau} \in \mathbb{R}^{H_{F} \times W_{F} \times 1}$ is the class agnostic center heatmap fetched from the $\bm{P}^{t-\tau}$ (as defined in \cref{sec:preliminaries}). Then, given $\bm{O}^{D}$ and $\bm{\bar{f}}^{t-\tau}$, the propagated feature is computed via a DCN as $\bm{\hat{f}}^{t-\tau}=DCN(\bm{O}^{D}, \bm{\bar{f}}^{t-\tau}) \in \mathbb{R}^{H_{F} \times W_{F} \times 64}$.

\myparagraph{Feature Enhancement:} When occlusion or motion blur occurs, objects could be missed by the detector. We propose to enhance $\bm{f}^{t}$ by aggregating the propagated feature $\bm{\hat{f}}^{t-\tau}$, on which the occluded and blurred objects may be visually legible. We denote the enhanced feature as $\bm{\tilde{f}}^{t-\tau}$, which is calculated by weighted summation as:
\vspace{-2mm}
\begin{equation}
\label{eqn:feat_agg}
\bm{\tilde{f}}^{t}_{q}=\bm{w}^{t}\circ\bm{f}^{t}_{q}+\sum_{\tau=1}^{T}\bm{w}^{t-\tau}\circ\bm{\hat{f}}^{t-\tau}_{q}, \quad q=1,2,...,64,
\vspace{-2mm}
\end{equation}
where $\bm{w}^{t} \in \mathbb{R}^{H_{F} \times W_{F} \times 1}$ is the adaptive weight at time $t$ and $\sum_{\tau=0}^{T}\bm{w}^{t-\tau}_{i,j}=1$. $T$ is the number of previous features used for aggregation. Similar to~\cite{liu2019learning}, $\bm{w}$ is predicted by two convolution layers followed by softmax function. We find that in experiment the weighted summation is slightly better than average summation. The enhanced feature $\bm{\tilde{f}}^{t}$ is then fed into the head networks to produce detection boxes and masks in the current frame. This can potentially recover missed objects and reduce false negatives, enabling complete tracklets and higher MOTA and IDF1 as in Tab.~\ref{tab:ablation1}.
\vspace{-3mm}

\subsection{Tracklet Generation}
\vspace{-1mm}
\label{subsec:tracklet_generation}
The overall architecture of TraDeS is shown in Fig.~\ref{fig:overview}. Based on the enhanced feature $\bm{\tilde{f}}^{t}$, TraDeS produces 2D and 3D boxes and instance masks by three different head networks. Afterwards, the generated detection and masks are connected to previous tracklets by our data association.

\myparagraph{Head Networks:} 
Each head network consists of several light-weight convolutions for yielding task-specific predictions. For 2D and 3D detection, we utilize the same head networks as in CenterNet~\cite{zhou2019objects}. For instance segmentation, we refer to the head network in CondInst~\cite{tian2020conditional}, which is an instance segmentation method also based on center points.

\myparagraph{Data Association:}
Given an enhanced detection or mask $d$ centered at location $(i,j)$, we perform a two-round data association as: \textbf{\emph{DA-Round (i)}} We first associate it with the closest unmatched detection at time $t-1$ within the area centered at $(i,j)+\bm{O}^{C}_{i,j}$ with radius $r$, where $r$ is the geometrical average of width and height of the detected box. Here, $\bm{O}^{C}_{i,j}$ only indicates the object tracking offsets between $\bm{I}^{t}$ and $\bm{I}^{t-1}$. \textbf{\emph{DA-Round (ii)}} If $d$ does not match any targets in the first round, we compute cosine similarities of its embedding $\bm{e}^{t}_{i,j}$ with all unmatched or history tracklet embeddings. $d$ will be assigned to a tracklet if their similarity is the highest and larger than a threshold, \emph{e.g.,} 0.3. \emph{DA-Round (ii)} is capable of long-term associating. In case $d$ fails to associate with any tacklets in the above two rounds, $d$ starts a new tracklet.

\myparagraph{TraDeS Loss:} The overall loss function of TraDeS is defined as $L = L_{CVA}+L_{det}+L_{mask}$, where $L_{det}$ is the 2D and 3D detection losses as in~\cite{zhou2019objects} and $L_{mask}$ is the instance segmentation loss as in~\cite{tian2020conditional}.

\begin{table*}[t]\centering\vspace{-3mm}
	% subfloat ############
	\subfloat[\textbf{Effectiveness of each proposed module}:  we evaluate the proposed CVA (\cref{subsec:CVA}), MFW (\cref{subsec:MFW}), and overall TraDeS (\cref{sec:trades}). ``Baseline+CVA+MFW'' is represented by ``TraDeS''.\label{tab:ablation1}]{
		\tablestyle{3pt}{1.05}
		\begin{tabular}{l|x{22}x{22}x{22}x{22}x{22}}	
			\rowcolor{mygray}
			Scheme  &MOTA$\uparrow$&IDF1$\uparrow$&IDS$\downarrow$&FN$\downarrow$&FP$\downarrow$ \\
			\shline
			CenterTrack\cite{CenterTrack} & 66.1&64.2&528&28.4\%&4.5\%\\
			\hline
			Baseline&64.8&59.5&1055&31.0\%&2.3 \% \\	
			Baseline+CVA&66.5&66.7&415&30.6\%&\bf 2.2\%\\
			Baseline+MFW&66.3&65.7&606&29.5\%&3.0\%\\
			TraDeS &\bf68.2&\bf 71.7&\bf 285&\bf 27.8\%&3.5\%\\
	\end{tabular}}\hspace{3mm}
\vspace{-7mm}
	% subfloat ############
	\subfloat[\textbf{CVA \emph{vs.} Common embedding}: 
	Common embedding loss $L_{CEembed}$ may downgrade detection performance, while our CVA learns an effective embedding without hurting detection. 
	As ``Baseline" does not have embedding, it only performs \emph{DA-Round(i)}. 
	\label{tab:ablation2}]{
		\tablestyle{3.5pt}{1.05}
		\begin{tabular}{ll|x{22}x{22}x{22}x{22}x{22}}
	\rowcolor{mygray}
	 \multicolumn{2}{c|}{Scheme} &MOTA$\uparrow$&IDF1$\uparrow$&IDS$\downarrow$&FN$\downarrow$&FP$\downarrow$ \\
	\shline
		&Baseline&64.8&59.5&1055&31.0\%&2.3\% \\
		\hline	
		\multirow{2}{*}{w/o \textbf{\emph{DA-Round (ii)}}}&+CE embedding&63.7&59.6&1099&32.1\%&\bf 2.2\% \\
		&+CVA&65.5&60.9&936&\bf30.6\%&\bf 2.2\%\\
		\hline
		\multirow{2}{*}{w/ \textbf{\emph{DA-Round (ii)}}}&+CE embedding&64.5&64.3&671&32.1\%&\bf 2.2\% \\
		&+CVA&\bf 66.5&\bf 66.7&\bf 415&\bf 30.6\%&\bf 2.2\%\\
		\end{tabular}}\hspace{3mm}
	\vspace{-3mm}
	\subfloat[\textbf{Motion cues:} In MFW, we evaluate different motion cues as the input of $\gamma(\cdot)$ to predict the DCN input offset $\bm{O}^{D}$. Ablations are based on baseline with CVA.\label{tab:ablation3}]{
		\tablestyle{3.5pt}{1.05}
		\begin{tabular}{ll|x{22}x{22}x{22}x{22}x{22}} 	
			\rowcolor{mygray}
	&Scheme&MOTA$\uparrow$&IDF1$\uparrow$&IDS$\downarrow$&FN$\downarrow$&FP$\downarrow$ \\
	\shline
	&Baseline+CVA&66.5&66.7&415&30.6\%&\bf 2.2\%\\
	\hline
	\multirow{2}{*}{TraDeS w/}
	%&$\bm{O}^{C}$ only&67.4&67.8&342&29.3\%&2.7\%\\
	&$\bm{f}^{t}-\bm{f}^{t-\tau}$ only &67.1&68.8&\bf 273&29.9\%&2.5\%\\
	%&$\bm{O}^{C}$ only&67.4&67.8&342&29.3\%&2.7\%\\
	&$\bm{f}^{t}-\bm{f}^{t-\tau}$ \& $\bm{O}^{C}$ &\bf68.2&\bf 71.7&285&\bf 27.8\%&3.5\%\\
	%\multicolumn{6}{c}{~}\\
	\end{tabular}}\hspace{6mm}
	% subfloat ############
	%\setlength{\tabcolsep}{4pt}
	 \hspace{-3mm}
	\subfloat[\textbf{Number of previous features}: We evaluate the MFW when aggregating different numbers of previous features. \label{tab:ablation4}]{
		\tablestyle{3pt}{1.05}
		\begin{tabular}{l|x{22}x{22}x{22}x{22}x{22}x{34}}
			\rowcolor{mygray}
	 Scheme&MOTA$\uparrow$&IDF1$\uparrow$&IDS$\downarrow$&FN$\downarrow$&FP$\downarrow$&Time(ms)$\downarrow$ \\
	\shline
	$T=1$&67.8&69.0&350&28.2\%&3.4\%&46\\
	$T=2$&\bf68.2&\bf 71.7&285&\bf 27.8\%&3.5\%&57\\
	$T=3$&67.5&69.9&\bf283&29.2\%&\bf2.8\%&70\\
	%\multicolumn{6}{c}{~}\\
	\end{tabular}}%\hspace{3mm}
	\vspace{4mm}
	% main caption
	\caption{\textbf{Ablation studies} on the MOT17 validation set. MOTA and IDF1 reflect the comprehensive tracking performance, while FN and FP reflect the detection performance. Lower FN means more missed objects are recovered. $\downarrow$ denotes lower is better. $\uparrow$ denotes higher is better.}
	\label{tab:ablations}
	\vspace{-3mm}
\end{table*}

\begin{figure*}
	\centering
	\includegraphics[width=1\linewidth]{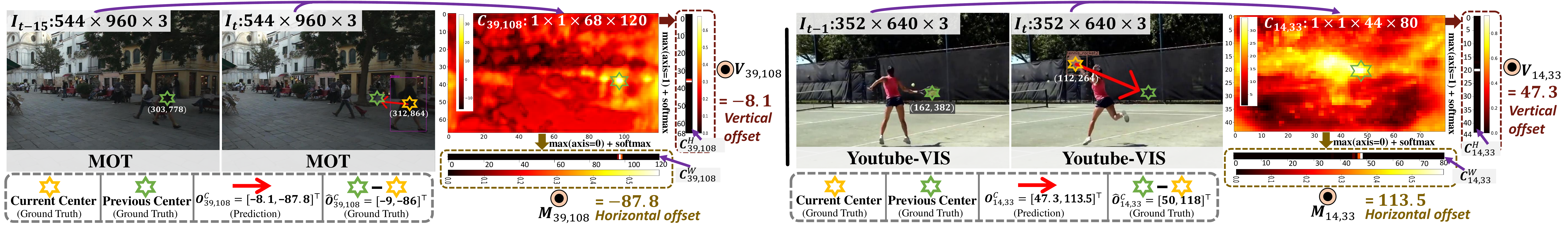}
	\caption{\textbf{CVA workflow visualization:} the cost volume map $\bm{C}$ and tracking offset $\bm{O}^{C}$ under low frame rate (left) and large motion (right).}
	\label{fig:ablation1}
	\vspace{-1.5mm}
\end{figure*}

\section{Experiments}
\label{sec:experiments}
\subsection{Datasets and Implementation Details}
\label{subsec:dataset_implement_details}
\myparagraph{MOT:} We conduct 2D object tracking experiments on the MOT16 and MOT17 datasets~\cite{milan1603mot16}, which have the same 7 training sequences and 7 test sequences but slightly different annotations. Frames are labeled at 25-30 FPS. For ablation study, we split the MOT17 training sequences into two halves and use one for training and the other for validation as in~\cite{CenterTrack}. \textbf{\emph{Metrics:}} We use common 2D MOT evaluation metrics~\cite{bernardin2008evaluating}: Multiple-Object Tracking Accuracy (MOTA), ID F1 Score (IDF1), the number of False Negatives (FN), False Positives (FP), times a trajectory is Fragmented (Frag), Identity Switches (IDS), and the percentage of Mostly Tracked Trajectories (MT) and Mostly Lost Trajectories (ML).

\myparagraph{nuScenes:} We conduct 3D object tracking experiments on the newly released nuScenes~\cite{caesar2020nuscenes}, containing 7 classes, 700 training sequences, 150 validation sequences,
and 150 test sequences. Videos are captured by 6 cameras of a moving car in a panoramic view and labeled at 2 FPS. Our TraDeS is a monocular tracker. \textbf{\emph{Metrics:}} nuScenes designs more robust metrics, AMOTA and AMOTP, which are computed by weighted averages of MOTA and MOTP across score thresholds from 0 to 1. For fair comparison, we also report IDS$_{A}$ that averages IDS in the same way. 

\myparagraph{MOTS:} MOTS~\cite{mots_dataset}, an instance segmentation tracking dataset, is derived from the MOT dataset. MOTS has 4 training sequences and 4 test sequences. \textbf{\emph{Metrics:}} The evaluation metrics are similar to those on MOT, which however are based on masks. Moreover, the MOTS adopts a Mask-based Soft Multi-Object Tracking Accuracy (sMOTSA).

\myparagraph{YouTube-VIS:} We also conduct instance segmentation tracking on YouTube-VIS~\cite{yang2019video}, which contains 2,883 videos labeled at 6 FPS, 131K instance masks, and 40 object classes. \textbf{\emph{Metrics:}} The YouTube-VIS adopts a mask tracklets based average precision (AP) for evaluation.

Compared to MOT and MOTS, nuScenes and YouTube-VIS are of low frame rate and large motion, because only key frames are labeled and cameras are moving. In our experiments, only labeled frames are used as input.

\vspace{-1mm}
\myparagraph{Implementation Details:} We adopt the same experimental settings as CenterTrack~\cite{CenterTrack}, such as backbone, image sizes, pretraining, score thresholds, etc. Specifically, we adopt the DLA-34~\cite{yu2018deep} as the backbone network $\phi(\cdot)$. Our method is optimized with 32 batches and learning rate (lr) $1.25e-4$ dropping by a factor of 10. For MOT and MOTS, TraDeS is trained for 70 epochs where lr drops at epoch 60 with image size of $544\times960$. For nuScenes, TraDeS is trained for 35 epochs where lr drops at epoch 30 with image size of $448\times800$. For YouTube-VIS, TraDeS is first pretrained on COCO instance segmentation~\cite{lin2014microsoft} following the static image training scheme in~\cite{CenterTrack} and then finetuned on YouTube-VIS for 16 epochs where lr drops at epoch 9. Image size is of $352\times640$. We test the runtime on a 2080Ti GPU. In Eq.~\ref{eqn:feat_agg}, we set $T=2$ by default for MOT and MOTS. We set $T=1$ for nuScenes and YouTube-VIS due to their low frame rate characteristic mentioned above. In training, we randomly select $T$ frames out of nearby $R_{t}$ frames, where $R_{t}$ is 10 for MOT and MOTS and 5 for nuScenes and YouTube-VIS. During inference, only previous $T$ consecutive frames are used. Ablation experiments are conducted on the MOT17 dataset. In ablations, all variants without the CVA module perform the \emph{DA-Round (i)} by predicting a tracking offset $\bm{O}^{B}$ as in the baseline tracker (\cref{sec:preliminaries}).

\subsection{Ablation Studies}
\vspace{-2mm}
\label{subsec:ablation_studies}
\myparagraph{Effectiveness of TraDeS:} As shown in Tab.~\ref{tab:ablation1}, we compare our proposed CVA (\cref{subsec:CVA}), MFW (\cref{subsec:MFW}), and TraDeS (\cref{sec:trades}) with our baseline tracker (\cref{sec:preliminaries}) and CenterTrack~\cite{CenterTrack}. \textbf{\emph{CVA:}} Compared to the baseline, the CVA achieves \emph{better tracking} by reducing 60\% IDS and improving 7.2 IDF1, validating the effect of our tracking offset, re-ID embedding, and the two-round data association. \textbf{\emph{MFW:}} For ablation, we directly add the MFW to the baseline tracker. Since the tracking offset $\bm{O}^{C}$ is unavailable in the baseline, we only use $\bm{f}^{t}-\bm{f}^{t-\tau}$ as motion cues to predict the DCN offset $\bm{O}^{D}$. Compared to the baseline, the MFW achieves \emph{better detection} by reducing 1.5\% FN, \emph{i.e.,} recovering more missed objects, though FP is slightly increased. Moreover, we observe that the MFW also reduces 43\% IDS and improves 6.2 IDF1. It validates that \emph{detection is the cornerstone for tracking performance}, where improved detection can yield more stable and consistent tracklets. \textbf{\emph{TraDeS:}} With the help of CVA, TraDeS reduces IDS from 606 to 285. Moreover, in TraDeS, the robust tracking offsets $\bm{O}^{C}$ from CVA guides the feature propagation in MFW, which significantly decreases FN from 29.5\% to 27.8\%. Better IDS and missed object recovery ($\downarrow$FN) together improve our comprehensive tracking performance, achieving 68.2 MOTA and 71.7 IDF1. TraDeS also achieves better results than the recent \emph{JDT} method CenterTrack~\cite{CenterTrack}.

\begin{table*}[!htb]
	\small
	\begin{center}
		\setlength{\tabcolsep}{3.35pt}
		\begin{tabular}{l|cc|c|ccccccccc}
			\shline
			\rowcolor{mygray}
			\multicolumn{13}{c}{\emph{MOT16 Test Set}}\\
			\hline
			\rowcolor{mygray}
			Method&Publication&Year&Joint&MOTA$\uparrow$& IDF1$\uparrow$&MT$\uparrow$&ML$\downarrow$&FP$\downarrow$&FN$\downarrow$&Frag$\downarrow$&IDS$\downarrow$&Time(ms)$\downarrow$\\
			
			\hline
			\hline
			SORT\cite{bewley2016simple}\textsuperscript{\emph{Online}} &ICIP&2016&& 59.8 & 53.8 & 25.4\% & 22.7\% &8,698 &63,245&1,835&1,423 & 17+D\\
			MCMOT-HDM\cite{lee2016multi}\textsuperscript{\emph{Offline}}&ECCV &2016& & 62.4 & 51.6 & 31.5\% & 24.2\% &9,855&57,257&1,318& 1,394 & 27+D\\
			POI\cite{yu2016poi}\textsuperscript{\emph{Online}}&ECCVW &2016& & 66.1 & 65.1 & 34.0\% & 20.8\% &5,061&55,914&3,093& 805 & 101+D\\
			DeepSORT\cite{wojke2017simple}\textsuperscript{\emph{Online}}&ICIP&2017&& 61.4 & 62.2 & 32.8\% & 18.2\% &12,852 &56,668&2,008&781 & 25+D\\
			VMaxx\cite{wan2018multi}\textsuperscript{\emph{Online}} &ICIP&2018&& 62.6 & 49.2 & 32.7\% & 21.1\% &10,604	&56,182&1,534& 1,389 & 154+D\\
			RAN\cite{fang2018recurrent}\textsuperscript{\emph{Online}} &WACV&2018&& 63.0 & 63.8 & 39.9\% & 22.1\% &13,663&53,248&1,251&482 &625+D\\
			TAP\cite{zhou2018online}\textsuperscript{\emph{Online}}&ICPR&2018&& 64.8 & 73.5 & 38.5\% & 21.6\% &12,980&50,635&1,048& 571 & 55+D\\
			\hline
			TubeTK\cite{tubetk}\textsuperscript{\emph{Offline}}&CVPR&2020&\Checkmark& 64.0 & \textcolor{blue}{\bf59.4} &33.5\% & \textcolor{red}{\bf 19.4 \%} &10,962&53,626&\textcolor{red}{\bf1,366}& \textcolor{red}{\bf1,117} & 1000\\
			JDE\cite{wang2019towards}\textsuperscript{\emph{Online}}&ECCV &2020&\Checkmark& 64.4 & 55.8 & \textcolor{blue}{\bf35.4\%} & \textcolor{blue}{\bf20.0\%} &-&-&-& 1,544 & 45\\
			CTracker\textsuperscript{\emph{Online}} &ECCV &2020&\Checkmark& \textcolor{blue}{\bf67.6} & 57.2 & 32.9\% & 23.1\% &\textcolor{blue}{\bf8,934}&\textcolor{blue}{\bf48,305}&3,112& 1,897 & 29\\
			TraDeS (Ours)\textsuperscript{\emph{Online}} &CVPR&2021&\Checkmark& \textcolor{red}{\bf 70.1} & \textcolor{red}{\bf 64.7} & \textcolor{red}{\bf 37.3 \%} & \textcolor{blue}{\bf 20.0 \%} &\textcolor{red}{\bf8,091}	&\textcolor{red}{\bf45,210}&\textcolor{blue}{\bf 1,575}& \textcolor{blue}{\bf 1,144} & 57\\
			\shline 
			
			\rowcolor{mygray}
			\multicolumn{13}{c}{\emph{MOT17 Test Set}}\\
			\hline
			\hline
			CenterTrack$^{\star}$\cite{CenterTrack}\textsuperscript{\emph{Online}} &ECCV&2020&\Checkmark& 67.8& 64.7 & 34.6\% & 24.6\% &18,498&160,332&6,102& 3,039 & 57\\
			TraDeS$^{\star}$ (Ours)\textsuperscript{\emph{Online}}&CVPR&2021&\Checkmark&68.9&67.2&35.0\%&22.7\%&19,701&152,622&6,033&3,147&57 \\
			\hline
			DAN\cite{sun2019deep}\textsuperscript{\emph{Online}} &TPAMI&2019&& 52.4 & 49.5 & 21.4\% & 30.7\% &25,423&234,592&14,797& 8,431 & 159+D\\
			Tracktor+CTdet\cite{tracktor}\textsuperscript{\emph{Online}}&ICCV&2019&&54.4&56.1&25.7\%&29.8\%&44,109&210,774&-&2,574&- \\
			\hline
			TubeTK\cite{tubetk}\textsuperscript{\emph{Offline}}&CVPR&2020 &\Checkmark& 63.0 & 58.6 & 31.2\% & \textcolor{red}{\bf19.9\%} &27,060&177,483&\textcolor{blue}{\bf 5,727}& 4,137 & 333\\
			CTracker\cite{CTacker}\textsuperscript{\emph{Online}}&ECCV&2020&\Checkmark& 66.6 & 57.4 & 32.2\% & 24.2\% &\textcolor{blue}{\bf 22,284}&160,491&9,114& 5,529 & 29\\
			CenterTrack\cite{CenterTrack}\textsuperscript{\emph{Online}}&ECCV&2020 &\Checkmark& \textcolor{blue}{\bf67.3} & \textcolor{blue}{\bf59.9} & \textcolor{blue}{\bf34.9\%} & 24.8\% &23,031&\textcolor{blue}{\bf158,676}&-& \textcolor{red}{\bf 2,898} & 57\\
			TraDeS (Ours)\textsuperscript{\emph{Online}} &CVPR&2021&\Checkmark& \textcolor{red}{\bf 69.1} & \textcolor{red}{\bf 63.9} & \textcolor{red}{\bf 36.4 \%} & \textcolor{blue}{\bf 21.5 \%} &\textcolor{red}{\bf 20,892}&\textcolor{red}{\bf 150,060}&\textcolor{red}{\bf 4,833}& \textcolor{blue}{\bf 3,555} & 57\\
			\hline
		\end{tabular}
	\end{center}
\vspace{-3mm}
\caption{\textbf{Results of 2D object tracking on the MOT test set under the private detection protocol.} ``Joint" indicates joint detection and tracking in a single model, \emph{i.e.,} no external detections. ``$\star$'' indicates that Track Re-birth~\cite{CenterTrack} is used. The top two results in the ``Joint" manner without Track Re-birth are highlighted in \textcolor{red}{\bf red} and \textcolor{blue}{\bf blue}, respectively. +D indicates the additional detection time~\cite{ren2015faster}.}
\label{tab:mot_sota}
\vspace{-3mm}
\end{table*}

\begin{figure*}
	\centering
	\includegraphics[width=1\linewidth]{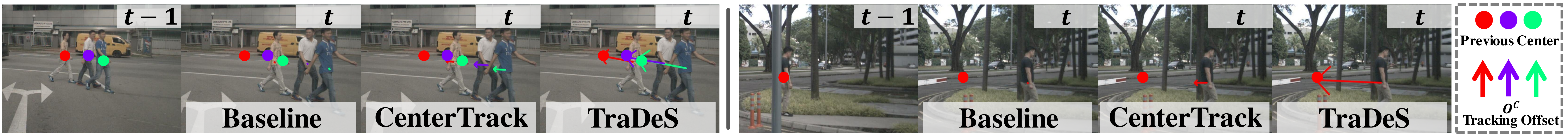}
	\caption{\textbf{Visualized $\bm{O}^{C}$ on nuScenes.} All models are only trained on MOT but tested on nuScenes, where nuScenes has much larger object motions than MOT. TraDeS successfully tracks objects with unseen large motion in training dataset, but baseline and CenterTrack fail.}
	\label{fig:ablation4}
	\vspace{-3mm}
\end{figure*}

\vspace{-1mm}
\myparagraph{Effectiveness of the CVA Module:} We study the two major characteristics of the proposed CVA module as mentioned in \cref{subsec:CVA}.  \textbf{\emph{(i):}} First, we add the re-ID embedding network $\sigma(\cdot)$ into the baseline tracker, which is supervised by a common re-ID loss, \emph{e.g.,} the cross-entropy loss $L_{CEembed}$ as in~\cite{wang2019towards,zhang2020fairmot}. We denote the learned embedding as CE embedding, which is used to perform our two-round data association. As shown in Tab.~\ref{tab:ablation2}, with {\emph{DA-Round (ii)}}, CE embedding helps baseline improve IDF1 and reduce IDS, as long-term data association is enabled by using the re-ID embedding to match history tracklets. However, we observe that CE embedding cannot improve MOTA as detection performance is degraded (+1.1\% FN). Next, we still add $\sigma(\cdot)$ into the baseline tracker, which however is supervised by our CVA module. Tab.~\ref{tab:ablation2} shows that our CVA module not only learns an effective re-ID embedding as CE embedding but also slightly improves detection performance, which clearly leads to a higher MOTA. We argue that this is because common re-ID loss only emphasizes intra-class variance, which may not be compatible with detection loss in joint training as indicted in~\cite{chen2018person}. In contrast, our proposed $L_{CVA}$ in Eq.~\ref{eqn:cva_loss} supervises the re-ID embedding via the cost volume and considers both intra-class and inter-class difference.
\textbf{\emph{(ii):}} We visualize the predicted cost volume map $\bm{C}$ and tracking offset $\bm{O}^{C}$ in Fig.~\ref{fig:ablation1}. The CVA accurately predicts the tracking offset for an object under low frame rate or large motion. Moreover, $\bm{O}^{C}$ even accurately tracks objects in a new dataset with unseen large motion in training as shown in Fig.~\ref{fig:ablation4}. Visualization of $\bm{O}^{C}$ on more samples are shown in Fig.~\ref{fig:final_results}. These examples indicate the CVA is able to predict tracking offsets for objects with a wide range of motion and provide robust motion cues.

\begin{figure}
	\centering
	\vspace{-3mm}
	\includegraphics[width=1\linewidth]{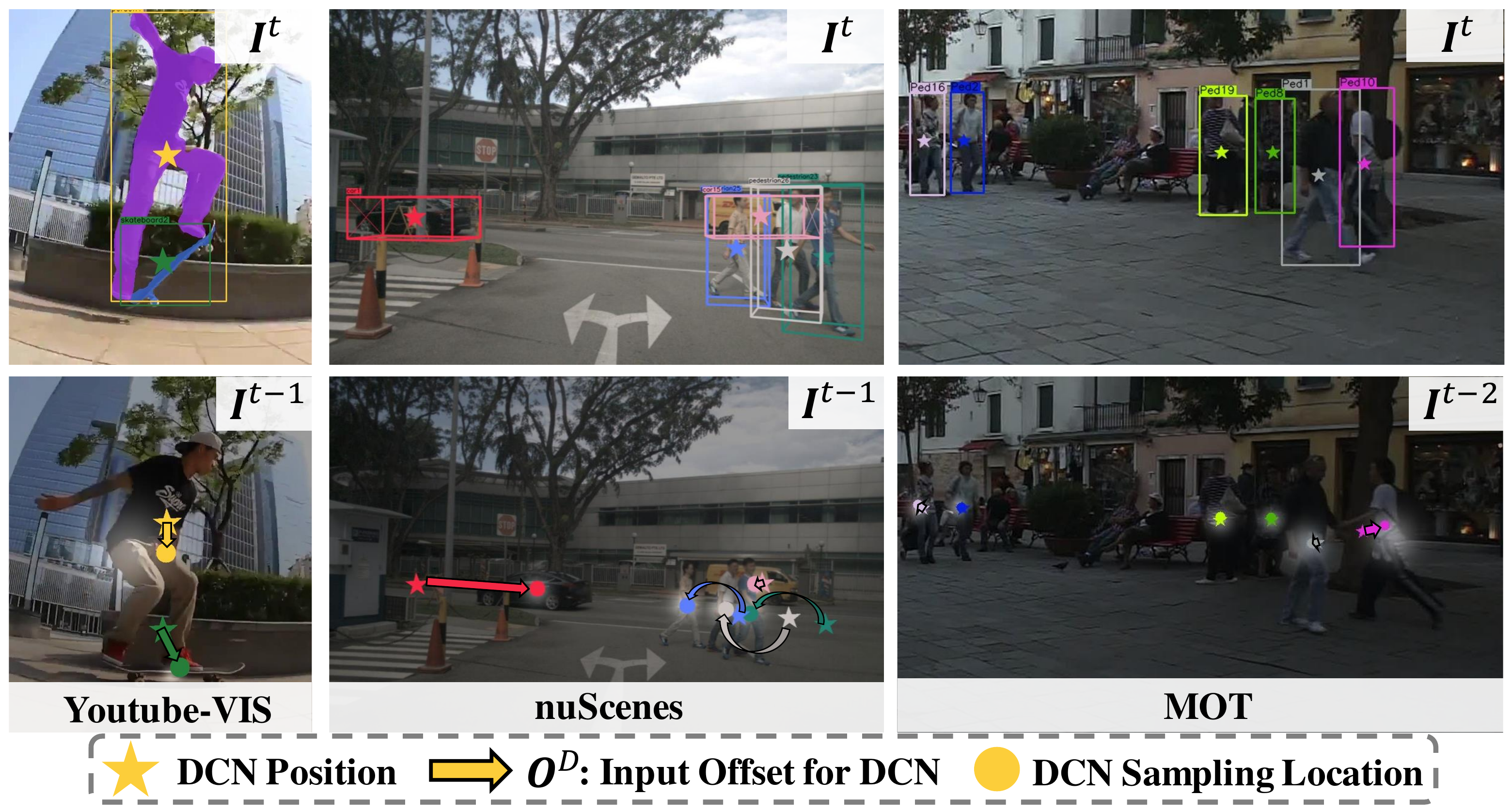}
	\caption{\textbf{Visualization of DCN input offset $\bm{O}^{D}$.} The DCN kernel at \protect\includegraphics[scale=0.08, trim=0 1.0cm 0 0]{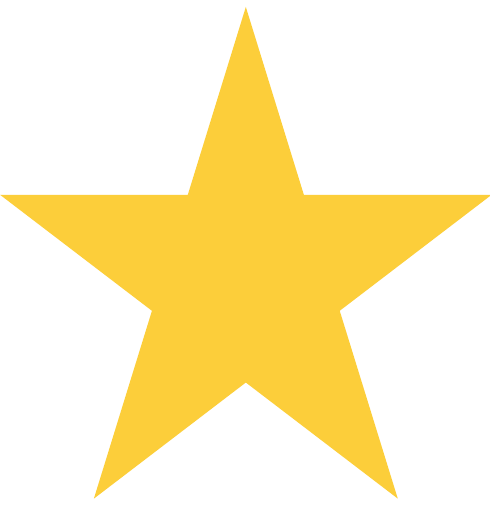} is translated by \protect\includegraphics[scale=0.1, trim=0 0.6cm 0 0]{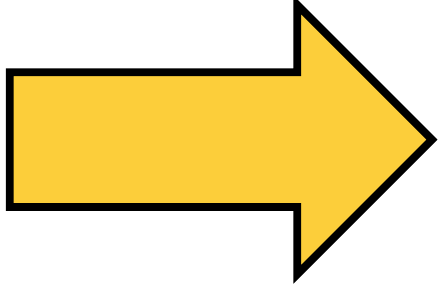} and samples the previous feature at \protect\includegraphics[scale=0.068, trim=0 1.0cm 0 0]{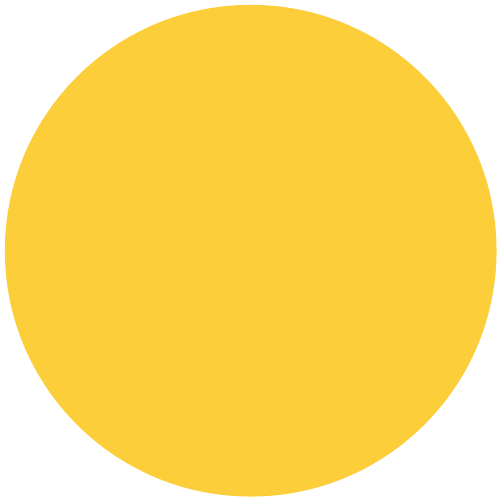}. For clear visualization, we only show the sampling center of the DCN kernel as depicted by \protect\includegraphics[scale=0.068, trim=0 1.0cm 0 0]{figures/markers/DCN_sampling_location/DCN_sampling_location_cropped.pdf} in $\bm{I}^{t-\tau}$. The previous image is highlighted by the previous class agnostic heatmap $\bm{P}_{agn}^{t-\tau}$.}
	\label{fig:ablation2}
	\vspace{-2mm}
\end{figure}

\myparagraph{Effectiveness of the MFW Module:} \textbf{\emph{DCN:}} In Tab.~\ref{tab:ablation3}, we use different motion clues to predict the DCN input offset $\bm{O}^{D}$. We find that the tracking offset $\bm{O}^{C}$ is the key to reduce FN and recover more missed objects. It validates that the proposed $\bm{O}^{C}$ is a robust tracking cue for guiding feature propagation and assisting detection. Moreover, we visualize the predicted $\bm{O}^{D}$ in Fig.~\ref{fig:ablation2}. The DCN successfully samples the center features at the previous frames even if the car in the middle image has dramatic displacements. \textbf{\emph{Number of Previous Features:}} As in Eq.~\ref{eqn:feat_agg}, the MFW aggregates the current feature with $T$ previous features. We evaluate the MFW with different $T$ as shown in Tab.~\ref{tab:ablation4}, and find that we achieve the best speed-accuracy trade-off when $T=2$.

\begin{table*}[!htb]
	\begin{center}	
		\setlength{\tabcolsep}{3.25pt}
		%\tablestyle{3.25pt}{1.0}
		\small
		\begin{tabular}{l|ccc|ccc|cccc }
			\shline
			\rowcolor{mygray}
			Classes&\multicolumn{3}{c|}{Car (57\% \tiny 58,317GTs \small)}&\multicolumn{3}{c|}{Pedestrian (25\% \tiny 25,423GTs \small)}&\multicolumn{4}{c}{All (100\% \tiny 101,897GTs \small)}\\
			\rowcolor{mygray}
			\emph{nuScenes Val set}&AMOTA$\uparrow$&AMOTP$\downarrow$&IDS$_{A}\downarrow$& AMOTA$\uparrow$&AMOTP$\downarrow$&IDS$_{A}\downarrow$&AMOTA$\uparrow$&AMOTP$\downarrow$&IDS$_{A}\downarrow$&Time \\ 
			\hline \hline
			Our Baseline&11.1&1.39&6,985&0.0&1.73&4,336&4.3&1.65&1,792&37ms \\	
			CenterTrack\cite{CenterTrack}&26.1&1.11&3,217&5.9&1.50&1,970&6.8&1.54&813&45ms \\		
			TraDeS (Ours)&\textcolor{red}{\bf29.6}&\textcolor{red}{\bf0.98} &\textcolor{red}{\bf3,035}&\textcolor{red}{\bf10.6}&\textcolor{red}{\bf1.42}&\textcolor{red}{\bf1,434} &\textcolor{red}{\bf11.8}&\textcolor{red}{\bf1.48}&\textcolor{red}{\bf699}&39ms \\		
			\shline
			\rowcolor{mygray}
			Classes&\multicolumn{3}{c|}{Car (57\% \tiny 68,518GTs \small)}&\multicolumn{3}{c|}{Pedestrian (28\% \tiny 34,010GTs \small)}&\multicolumn{4}{c}{All (100\% \tiny 119,565GTs \small)} \\
			\rowcolor{mygray}
			\emph{nuScenes Test set}&AMOTA$\uparrow$&AMOTP$\downarrow$&IDS$_{A}\downarrow$  &AMOTA$\uparrow$&AMOTP$\downarrow$&IDS$_{A}\downarrow$&AMOTA$\uparrow$&AMOTP$\downarrow$&IDS$_{A}\downarrow$&Time\\ 
			\hline \hline
			Our Baseline&6.2&1.47&9,450&0.0&1.70&5,191&1.0&1.66&2,252&37ms \\
			Mapillary\cite{simonelli2019disentangling}+AB3D\cite{Weng2019_3dmot}&12.5&1.61&-&0.0&1.87&-&1.8&1.80&-&- \\
			PointPillars\cite{lang2019pointpillars}+AB3D\cite{Weng2019_3dmot}&9.4&1.40&-&3.9&1.68&-&2.9&1.70&-&- \\
			CenterTrack\cite{CenterTrack}&20.2&1.19&-&3.0&1.50&-&4.6&1.54&-&45ms \\		
			TraDeS (Ours)&\textcolor{red}{\bf23.2}&\textcolor{red}{\bf1.07}&\textcolor{red}{\bf4,293}&\textcolor{red}{\bf9.9}&\textcolor{red}{\bf1.38}&\textcolor{red}{\bf1,979}&\textcolor{red}{\bf5.9}&\textcolor{red}{\bf1.49}&\textcolor{red}{\bf964} &39ms \\		
			\hline
		\end{tabular}
	\vspace{-2mm}	
	\end{center}
	\caption{\textbf{Results of 3D object tracking on the nuScenes dataset.} We compare with the state-of-the-art monocular 3D tracking methods. We mainly assess the major classes: car and pedestrian. We also list ``All'' for reference, which is the average among all the 7 classes.}
	\label{tab:nuscenes_sota}
	\vspace{-2mm}
\end{table*}

\begin{table*}[!htb]
	\begin{center}	
		\setlength{\tabcolsep}{1.5pt}
		\small
		\begin{tabular}{l|cc|ccccccccccc }
			\shline
			\rowcolor{mygray}
			Method&Publication&Year&sMOTSA $\uparrow$&IDF1 $\uparrow$&MOTSA $\uparrow$&MOTSP $\uparrow$&MODSA $\uparrow$&MT$\uparrow$ & ML$\downarrow$ &FP$\downarrow$&FN$\downarrow$&IDS $\downarrow$&Time\\ 
			\hline
			\hline
			TrackR-CNN~\cite{mots_dataset} &CVPR&2019&40.6&42.4&55.2&76.1&56.9&38.7\%&21.6\%&\textcolor{red}{\bf 1,261}&12,641&567&500ms\\
			TraDeS (Ours)&CVPR&2021&\textcolor{red}{\bf 50.8}&\textcolor{red}{\bf 58.7}&\textcolor{red}{\bf 65.5} &\textcolor{red}{\bf 79.5}&\textcolor{red}{\bf 67.0}&\textcolor{red}{\bf 49.4\%}&\textcolor{red}{\bf 18.3\%}&1,474&\textcolor{red}{\bf 9,169}&\textcolor{red}{\bf 492}&87ms\\
			\hline
		\end{tabular}
	\vspace{-2mm}
	\end{center}
	\caption{\textbf{Results of instance segmentation tracking on the MOTS test set.}}
	\label{tab:mots_sota}
	\vspace{-3mm}
\end{table*}

\begin{figure*}[h!]
	\centering
	\vspace{-1mm}
	\includegraphics[width=1\linewidth]{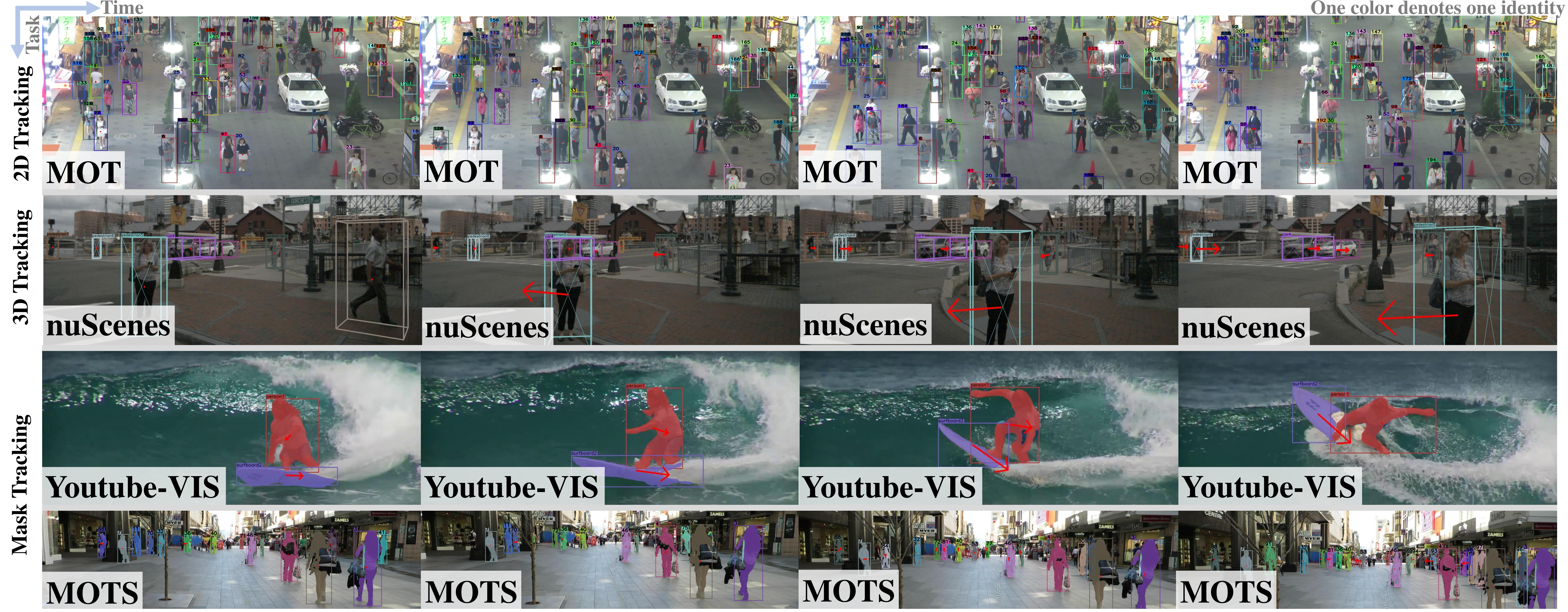}
	\vspace{-4mm}
	\caption{\textbf{Visualization that TraDeS tracks objects on three tasks.} Red arrow is the tracking offset $\bm{O}^{C}$ \emph{w.r.t.} the previous frame $\bm{I}^{t-1}$.}
	\label{fig:final_results}
	\vspace{-3mm}
\end{figure*}

\begin{table}[h]
	\vspace{-2mm}
	\begin{center}	
		\small
		\setlength{\tabcolsep}{3pt}
		\begin{tabular}{l|c|ccc}
			\shline
			\rowcolor{mygray}
			Method&Publication&AP&AP$_{50}$&AP$_{75}$\\ 
			\hline \hline
			OSMN(mask propagation)\cite{Yang_2018_CVPR}&CVPR'18&23.4&36.5&25.7\\
			FEELVOS\cite{voigtlaender2019feelvos}&CVPR'19& 26.9&42.0&29.7\\
			OSMN(track-by-detect)\cite{Yang_2018_CVPR}&CVPR'18&27.5&45.1&29.1\\
			MaskTrack R-CNN\cite{yang2019video}&ICCV'19&30.3&51.1&32.6\\
			SipMask\cite{cao2020sipmask}&ECCV'20&32.5&\textcolor{red}{\bf53.0}&\textcolor{red}{\bf33.3}\\
			\hline
			Our Baseline&&26.4&43.2&26.8\\
			TraDeS (Ours)&CVPR'21&\textcolor{red}{\bf32.6}&52.6&32.8\\
			\hline
		\end{tabular}
	\end{center}
	\vspace{-2mm}
	\caption{\textbf{Results of instance segmentation tracking on the YouTube-VIS validation set.}}
	\label{tab:youtube_vis_sota}
	\vspace{-3mm}
\end{table}

\subsection{Benchmark Evaluations}
\label{subsec:sota}
\vspace{-2mm}
\myparagraph{MOT:} As shown in Tab.~\ref{tab:mot_sota}, we compare the proposed TraDeS tracker with the state-of-the-art 2D trackers on the MOT16 and MOT17 test sets. Our TraDeS tracker outperforms the second best tracker by \emph{2.5} MOTA and \emph{1.8} MOTA on MOT16 and MOT17, respectively, running at \emph{15 FPS}. Compared to joint detection and tracking algorithms, we achieve the best results on most metrics, \emph{e.g.,} MOTA, IDF1, MT, FN, etc. 

\myparagraph{nuScenes:} As shown in Tab.~\ref{tab:nuscenes_sota}, we compare TraDeS with the state-of-the-art monocular 3D trackers on nuScenes. There exists extreme class imbalance in nuScenes dataset, \emph{e.g.,} car and pedestrian has over 82\% data. Since class imbalance is not our focus, we mainly evaluate on major classes: car and pedestrian. Tab.~\ref{tab:nuscenes_sota} shows that the TraDeS tracker outperforms other monocular trackers by a large margin on all metrics.

\myparagraph{MOTS:} As shown in Tab.~\ref{tab:mots_sota}, we compare TraDeS with the recent instance segmentation tracker TrackR-CNN on the MOTS test set. TrackR-CNN is based on Mask R-CNN~\cite{he2017mask} and also temporally enhances object features. The TraDeS tracker outperforms TrackR-CNN by a large margin in terms of both accuracy and speed.

\myparagraph{YouTube-VIS:} As shown in Tab.~\ref{tab:youtube_vis_sota}, TraDeS notably improves AP by 6.2 over the baseline. TraDeS achieves competitive performance compared to other state-of-the-art instance segmentation trackers. We observe that TraDeS outperforms the baseline tracker by a large margin on both nuScenes and YouTube-VIS. We argue that this is because the baseline cannot well predict the tracking offset $\bm{O}^{B}$ with a single image in case these datasets are of low frame rate and large motion.

\vspace{-2mm}
\section{Conclusion}
\vspace{-2mm}
This work presents a novel online joint detection and tracking model, TraDeS, focusing on exploiting tracking cues to help detection and in return benefit tracking. TraDeS is equipped with two proposed modules, CVA and MFW. The CVA learns a dedicatedly designed re-ID embedding and models object motions via a 4d cost volume. While the MFW takes the motions from CVA as the cues to propagate previous object features to enhance the current detection or segmentation. Exhaustive experiments and ablations on 2D tracking, 3D tracking and instance segmentation tracking validate both effectiveness and superiority of our approach.

\myparagraph{Acknowledgement.} This work is supported in part by a gift grant from Horizon Robotics and National Science Foundation Grant CNS1951952. We thank Sijia Chen and Li Huang from Horizon Robotics for helpful discussion.

{\small
\bibliographystyle{ieee_fullname}
\bibliography{TraDeS}

\begin{thebibliography}{10}\itemsep=-1pt

\bibitem{tracktor}
Philipp Bergmann, Tim Meinhardt, and Laura Leal-Taixe.
\newblock Tracking without bells and whistles.
\newblock In {\em ICCV}, 2019.

\bibitem{bernardin2008evaluating}
Keni Bernardin and Rainer Stiefelhagen.
\newblock Evaluating multiple object tracking performance: the clear mot
  metrics.
\newblock {\em EURASIP Journal on Image and Video Processing}, 2008.

\bibitem{bertasius2019learning}
Gedas Bertasius, Christoph Feichtenhofer, Du Tran, Jianbo Shi, and Lorenzo
  Torresani.
\newblock Learning temporal pose estimation from sparsely-labeled videos.
\newblock In {\em NeurIPS}, 2019.

\bibitem{bertasius2018object}
Gedas Bertasius, Lorenzo Torresani, and Jianbo Shi.
\newblock Object detection in video with spatiotemporal sampling networks.
\newblock In {\em ECCV}, 2018.

\bibitem{bewley2016simple}
Alex Bewley, Zongyuan Ge, Lionel Ott, Fabio Ramos, and Ben Upcroft.
\newblock Simple online and realtime tracking.
\newblock In {\em ICIP}, 2016.

\bibitem{braso2020learning}
Guillem Bras{\'o} and Laura Leal-Taix{\'e}.
\newblock Learning a neural solver for multiple object tracking.
\newblock In {\em CVPR}, 2020.

\bibitem{caesar2020nuscenes}
Holger Caesar, Varun Bankiti, Alex~H Lang, Sourabh Vora, Venice~Erin Liong,
  Qiang Xu, Anush Krishnan, Yu Pan, Giancarlo Baldan, and Oscar Beijbom.
\newblock nuscenes: A multimodal dataset for autonomous driving.
\newblock In {\em CVPR}, 2020.

\bibitem{cao2020sipmask}
Jiale Cao, Rao~Muhammad Anwer, Hisham Cholakkal, Fahad~Shahbaz Khan, Yanwei
  Pang, and Ling Shao.
\newblock Sipmask: Spatial information preservation for fast image and video
  instance segmentation.
\newblock {\em ECCV}, 2020.

\bibitem{chen2018person}
Di Chen, Shanshan Zhang, Wanli Ouyang, Jian Yang, and Ying Tai.
\newblock Person search via a mask-guided two-stream cnn model.
\newblock In {\em ECCV}, 2018.

\bibitem{chen2016full}
Qifeng Chen and Vladlen Koltun.
\newblock Full flow: Optical flow estimation by global optimization over
  regular grids.
\newblock In {\em CVPR}, 2016.

\bibitem{collins1996space}
Robert~T Collins.
\newblock A space-sweep approach to true multi-image matching.
\newblock In {\em CVPR}, 1996.

\bibitem{dai2017deformable}
Jifeng Dai, Haozhi Qi, Yuwen Xiong, Yi Li, Guodong Zhang, Han Hu, and Yichen
  Wei.
\newblock Deformable convolutional networks.
\newblock In {\em ICCV}, 2017.

\bibitem{deng2020single}
Jiajun Deng, Yingwei Pan, Ting Yao, Wengang Zhou, Houqiang Li, and Tao Mei.
\newblock Single shot video object detector.
\newblock {\em TMM}, 2020.

\bibitem{fang2018recurrent}
Kuan Fang, Yu Xiang, Xiaocheng Li, and Silvio Savarese.
\newblock Recurrent autoregressive networks for online multi-object tracking.
\newblock In {\em WACV}, 2018.

\bibitem{feichtenhofer2017detect}
Christoph Feichtenhofer, Axel Pinz, and Andrew Zisserman.
\newblock Detect to track and track to detect.
\newblock In {\em ICCV}, 2017.

\bibitem{felzenszwalb2009object}
Pedro~F Felzenszwalb, Ross~B Girshick, David McAllester, and Deva Ramanan.
\newblock Object detection with discriminatively trained part-based models.
\newblock {\em TPAMI}, 2009.

\bibitem{he2017mask}
Kaiming He, Georgia Gkioxari, Piotr Doll{\'a}r, and Ross Girshick.
\newblock Mask r-cnn.
\newblock In {\em ICCV}, 2017.

\bibitem{im2019dpsnet}
Sunghoon Im, Hae-Gon Jeon, Stephen Lin, and In~So Kweon.
\newblock Dpsnet: End-to-end deep plane sweep stereo.
\newblock 2019.

\bibitem{kim2015multiple}
Chanho Kim, Fuxin Li, Arridhana Ciptadi, and James~M Rehg.
\newblock Multiple hypothesis tracking revisited.
\newblock In {\em ICCV}, 2015.

\bibitem{lang2019pointpillars}
Alex~H Lang, Sourabh Vora, Holger Caesar, Lubing Zhou, Jiong Yang, and Oscar
  Beijbom.
\newblock Pointpillars: Fast encoders for object detection from point clouds.
\newblock In {\em CVPR}, 2019.

\bibitem{lee2016multi}
Byungjae Lee, Enkhbayar Erdenee, Songguo Jin, Mi~Young Nam, Young~Giu Jung, and
  Phill~Kyu Rhee.
\newblock Multi-class multi-object tracking using changing point detection.
\newblock In {\em ECCV}, 2016.

\bibitem{lin2017focal}
Tsung-Yi Lin, Priya Goyal, Ross Girshick, Kaiming He, and Piotr Doll{\'a}r.
\newblock Focal loss for dense object detection.
\newblock In {\em ICCV}, 2017.

\bibitem{lin2014microsoft}
Tsung-Yi Lin, Michael Maire, Serge Belongie, James Hays, Pietro Perona, Deva
  Ramanan, Piotr Doll{\'a}r, and C~Lawrence Zitnick.
\newblock Microsoft coco: Common objects in context.
\newblock In {\em ECCV}, 2014.

\bibitem{liu2019learning}
Songtao Liu, Di Huang, and Yunhong Wang.
\newblock Learning spatial fusion for single-shot object detection.
\newblock {\em arXiv preprint arXiv:1911.09516}, 2019.

\bibitem{lu2020retinatrack}
Zhichao Lu, Vivek Rathod, Ronny Votel, and Jonathan Huang.
\newblock Retinatrack: Online single stage joint detection and tracking.
\newblock In {\em CVPR}, 2020.

\bibitem{meinhardt2021trackformer}
Tim Meinhardt, Alexander Kirillov, Laura Leal-Taixe, and Christoph
  Feichtenhofer.
\newblock Trackformer: Multi-object tracking with transformers.
\newblock {\em arXiv preprint arXiv:2101.02702}, 2021.

\bibitem{milan1603mot16}
Anton Milan, Laura Leal-Taix{\'e}, Ian Reid, Stefan Roth, and Konrad Schindler.
\newblock Mot16: A benchmark for multi-object tracking.
\newblock {\em arXiv preprint arXiv:1603.00831}.

\bibitem{tubetk}
Bo Pang, Yizhuo Li, Yifan Zhang, Muchen Li, and Cewu Lu.
\newblock Tubetk: Adopting tubes to track multi-object in a one-step training
  model.
\newblock In {\em CVPR}, 2020.

\bibitem{CTacker}
Jinlong Peng, Changan Wang, Fangbin Wan, Yang Wu, Yabiao Wang, Ying Tai,
  Chengjie Wang, Jilin Li, Feiyue Huang, and Yanwei Fu.
\newblock Chained-tracker: Chaining paired attentive regression results for
  end-to-end joint multiple-object detection and tracking.
\newblock In {\em ECCV}, 2020.

\bibitem{Porzi_2020_CVPR}
Lorenzo Porzi, Markus Hofinger, Idoia Ruiz, Joan Serrat, Samuel~Rota Bulo, and
  Peter Kontschieder.
\newblock Learning multi-object tracking and segmentation from automatic
  annotations.
\newblock In {\em CVPR}, 2020.

\bibitem{ren2015faster}
Shaoqing Ren, Kaiming He, Ross Girshick, and Jian Sun.
\newblock Faster r-cnn: Towards real-time object detection with region proposal
  networks.
\newblock In {\em NeurIPS}, 2015.

\bibitem{schroff2015facenet}
Florian Schroff, Dmitry Kalenichenko, and James Philbin.
\newblock Facenet: A unified embedding for face recognition and clustering.
\newblock In {\em CVPR}, 2015.

\bibitem{schulter2017deep}
Samuel Schulter, Paul Vernaza, Wongun Choi, and Manmohan Chandraker.
\newblock Deep network flow for multi-object tracking.
\newblock In {\em CVPR}, 2017.

\bibitem{simonelli2019disentangling}
Andrea Simonelli, Samuel Rota~Rota Bul{\`o}, Lorenzo Porzi, Manuel
  L{\'o}pez-Antequera, and Peter Kontschieder.
\newblock Disentangling monocular 3d object detection.
\newblock In {\em ICCV}, 2019.

\bibitem{sun2018pwc}
Deqing Sun, Xiaodong Yang, Ming-Yu Liu, and Jan Kautz.
\newblock Pwc-net: Cnns for optical flow using pyramid, warping, and cost
  volume.
\newblock In {\em CVPR}, 2018.

\bibitem{sun2020transtrack}
Peize Sun, Yi Jiang, Rufeng Zhang, Enze Xie, Jinkun Cao, Xinting Hu, Tao Kong,
  Zehuan Yuan, Changhu Wang, and Ping Luo.
\newblock Transtrack: Multiple-object tracking with transformer.
\newblock {\em arXiv preprint arXiv:2012.15460}, 2020.

\bibitem{sun2019deep}
Shijie Sun, Naveed Akhtar, HuanSheng Song, Ajmal~S Mian, and Mubarak Shah.
\newblock Deep affinity network for multiple object tracking.
\newblock {\em IEEE TPAMI}, 2019.

\bibitem{tang2017multiple}
Siyu Tang, Mykhaylo Andriluka, Bjoern Andres, and Bernt Schiele.
\newblock Multiple people tracking by lifted multicut and person
  re-identification.
\newblock In {\em CVPR}, 2017.

\bibitem{tian2020conditional}
Zhi Tian, Chunhua Shen, and Hao Chen.
\newblock Conditional convolutions for instance segmentation.
\newblock In {\em ECCV}, 2020.

\bibitem{voigtlaender2019feelvos}
Paul Voigtlaender, Yuning Chai, Florian Schroff, Hartwig Adam, Bastian Leibe,
  and Liang-Chieh Chen.
\newblock Feelvos: Fast end-to-end embedding learning for video object
  segmentation.
\newblock In {\em CVPR}, 2019.

\bibitem{mots_dataset}
Paul Voigtlaender, Michael Krause, Aljosa Osep, Jonathon Luiten, Berin
  Balachandar~Gnana Sekar, Andreas Geiger, and Bastian Leibe.
\newblock Mots: Multi-object tracking and segmentation.
\newblock In {\em CVPR}, 2019.

\bibitem{wan2018multi}
Xingyu Wan, Jinjun Wang, Zhifeng Kong, Qing Zhao, and Shunming Deng.
\newblock Multi-object tracking using online metric learning with long
  short-term memory.
\newblock In {\em ICIP}, 2018.

\bibitem{wang2020combining}
Manchen Wang, Joseph Tighe, and Davide Modolo.
\newblock Combining detection and tracking for human pose estimation in videos.
\newblock In {\em CVPR}, 2020.

\bibitem{wang2020joint}
Yongxin Wang, Kris Kitani, and Xinshuo Weng.
\newblock Joint object detection and multi-object tracking with graph neural
  networks.
\newblock {\em arXiv preprint arXiv:2006.13164}, 2020.

\bibitem{wang2019towards}
Zhongdao Wang, Liang Zheng, Yixuan Liu, and Shengjin Wang.
\newblock Towards real-time multi-object tracking.
\newblock In {\em ECCV}, 2020.

\bibitem{wen2014multiple}
Longyin Wen, Wenbo Li, Junjie Yan, Zhen Lei, Dong Yi, and Stan~Z Li.
\newblock Multiple target tracking based on undirected hierarchical relation
  hypergraph.
\newblock In {\em CVPR}, 2014.

\bibitem{Weng2019_3dmot}
Xinshuo Weng and Kris Kitani.
\newblock 3d multi-object tracking: A baseline and new evaluation metrics.
\newblock In {\em IROS}, 2020.

\bibitem{Weng_2020_CVPR}
Xinshuo Weng, Yongxin Wang, Yunze Man, and Kris~M. Kitani.
\newblock Gnn3dmot: Graph neural network for 3d multi-object tracking with
  2d-3d multi-feature learning.
\newblock In {\em CVPR}, 2020.

\bibitem{wojke2017simple}
Nicolai Wojke, Alex Bewley, and Dietrich Paulus.
\newblock Simple online and realtime tracking with a deep association metric.
\newblock In {\em ICIP}, 2017.

\bibitem{wu2020temporal}
Jialian Wu, Chunluan Zhou, Ming Yang, Qian Zhang, Yuan Li, and Junsong Yuan.
\newblock Temporal-context enhanced detection of heavily occluded pedestrians.
\newblock In {\em CVPR}, 2020.

\bibitem{xiao2017joint}
Tong Xiao, Shuang Li, Bochao Wang, Liang Lin, and Xiaogang Wang.
\newblock Joint detection and identification feature learning for person
  search.
\newblock In {\em CVPR}, 2017.

\bibitem{xu2019spatial}
Jiarui Xu, Yue Cao, Zheng Zhang, and Han Hu.
\newblock Spatial-temporal relation networks for multi-object tracking.
\newblock In {\em ICCV}, 2019.

\bibitem{xu2017accurate}
Jia Xu, Ren{\'e} Ranftl, and Vladlen Koltun.
\newblock Accurate optical flow via direct cost volume processing.
\newblock In {\em CVPR}, 2017.

\bibitem{xu2020segment}
Zhenbo Xu, Wei Zhang, Xiao Tan, Wei Yang, Huan Huang, Shilei Wen, Errui Ding,
  and Liusheng Huang.
\newblock Segment as points for efficient online multi-object tracking and
  segmentation.
\newblock In {\em ECCV}, 2020.

\bibitem{yang2020cost}
Jiayu Yang, Wei Mao, Jose~M Alvarez, and Miaomiao Liu.
\newblock Cost volume pyramid based depth inference for multi-view stereo.
\newblock In {\em CVPR}, 2020.

\bibitem{yang2019video}
Linjie Yang, Yuchen Fan, and Ning Xu.
\newblock Video instance segmentation.
\newblock In {\em ICCV}, 2019.

\bibitem{Yang_2018_CVPR}
Linjie Yang, Yanran Wang, Xuehan Xiong, Jianchao Yang, and Aggelos~K.
  Katsaggelos.
\newblock Efficient video object segmentation via network modulation.
\newblock In {\em CVPR}, 2018.

\bibitem{yin2020unified}
Junbo Yin, Wenguan Wang, Qinghao Meng, Ruigang Yang, and Jianbing Shen.
\newblock A unified object motion and affinity model for online multi-object
  tracking.
\newblock In {\em CVPR}, 2020.

\bibitem{yu2016poi}
Fengwei Yu, Wenbo Li, Quanquan Li, Yu Liu, Xiaohua Shi, and Junjie Yan.
\newblock Poi: Multiple object tracking with high performance detection and
  appearance feature.
\newblock In {\em ECCV Workshops}, 2016.

\bibitem{yu2018deep}
Fisher Yu, Dequan Wang, Evan Shelhamer, and Trevor Darrell.
\newblock Deep layer aggregation.
\newblock In {\em CVPR}, 2018.

\bibitem{zhang2020fairmot}
Yifu Zhang, Chunyu Wang, Xinggang Wang, Wenjun Zeng, and Wenyu Liu.
\newblock Fairmot: On the fairness of detection and re-identification in
  multiple object tracking.
\newblock {\em arXiv preprint arXiv:2004.01888}, 2020.

\bibitem{zhang2018integrated}
Zheng Zhang, Dazhi Cheng, Xizhou Zhu, Stephen Lin, and Jifeng Dai.
\newblock Integrated object detection and tracking with tracklet-conditioned
  detection.
\newblock {\em arXiv preprint arXiv:1811.11167}, 2018.

\bibitem{CenterTrack}
Xingyi Zhou, Vladlen Koltun, and Philipp Kr{\"a}henb{\"u}hl.
\newblock Tracking objects as points.
\newblock In {\em ECCV}, 2020.

\bibitem{zhou2019objects}
Xingyi Zhou, Dequan Wang, and Philipp Kr{\"a}henb{\"u}hl.
\newblock Objects as points.
\newblock {\em arXiv preprint arXiv:1904.07850}, 2019.

\bibitem{zhou2018online}
Zongwei Zhou, Junliang Xing, Mengdan Zhang, and Weiming Hu.
\newblock Online multi-target tracking with tensor-based high-order graph
  matching.
\newblock In {\em ICPR}, 2018.

\bibitem{zhu2018online}
Ji Zhu, Hua Yang, Nian Liu, Minyoung Kim, Wenjun Zhang, and Ming-Hsuan Yang.
\newblock Online multi-object tracking with dual matching attention networks.
\newblock In {\em ECCV}, 2018.

\end{thebibliography}
}
\end{document}